# Map Memorization and Forgetting in the IARA Autonomous Car*

Thomas Teixeira, Filipe Mutz, Vinicius Brito, Lucas Veronese, Claudine Badue, Thiago Oliveira-Santos, Alberto F. De Souza, *Senior Member, IEEE*

***Abstract*—In this work, we present a novel strategy to correct imperfections in occupancy grid maps called map decay. The objective of map decay is to correct invalid occupancy probabilities of map cells that are unobservable by sensors. The strategy was inspired by an analogy between the memory architecture believed to exist in the brain and the maps maintained by an autonomous vehicle. It consists in merging sensory information obtained in runtime (online) with *a priori* data from a high-precision map constructed offline. In map decay, cells observed by sensors are updated using traditional occupancy grid mapping techniques and unobserved cells are adjusted so that their occupancy probabilities tend to the values found in the offline map. This strategy is grounded in the idea that the most precise information available about an unobservable cell is the value found in the high-precision offline map. Map decay was successfully tested and still in use in the IARA autonomous vehicle from Universidade Federal do Espírito Santo.**

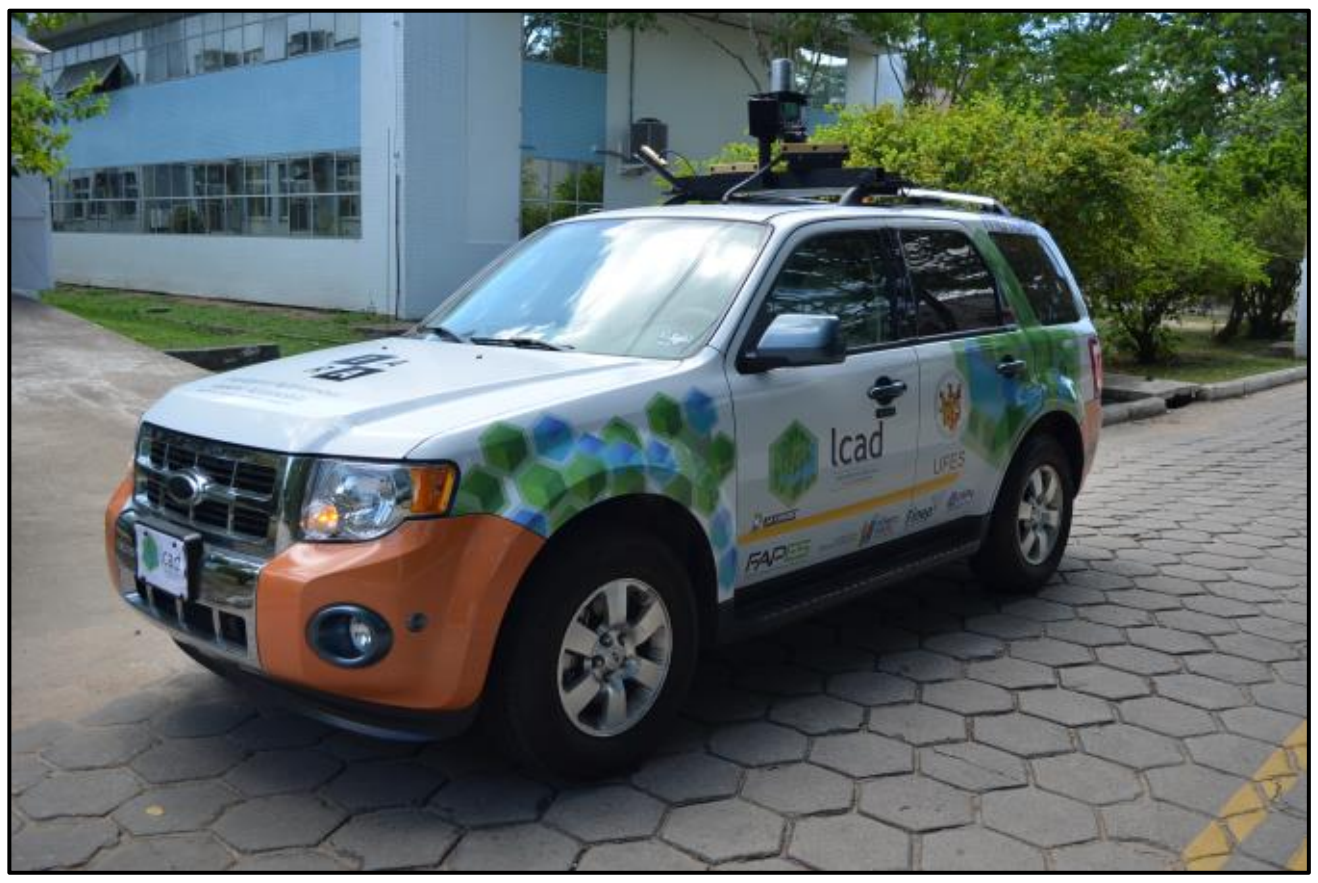

Figure 1. IARA autonomous vehicle from Universidade Federal do Espírito Santo. IARA is an achronym for Intelligent and Autonomous Robotic Automobile.

## I. Introduction

The brain has a key role in the human capacity of operating in highly dynamic and complex environments, and solving general purpose problems. The idea of giving these abilities to artificial entities by reproducing brain's cognitive processes always fascinated researchers. In several works the brain and its processes inspired algorithms and new strategies to solve problems. Milford and Wyeth, and posteriorly Ball et al. developed an artificial model of the rat's hippocampus, the region of the brain responsible for place and experience representation [1], [2]. This model was successfully used in Simultaneous Localization and Mapping (SLAM) applications. Rivest et al. analyzed the brain's dopaminergic pathways, i.e. the neural pathways associated with reward-motivated behavior, to propose new reinforcement learning algorithms [3]. Berger et al. presented a novel visual tracking technique inspired by the brain's regions related to saccadic eye movements [4]. The brain's visual cortex also inspired several other robotics applications in cognitive map building and scene understanding [5], self-motion estimation [6], and feature extraction and place recognition [7].

In this work, we also search for inspiration in the brain to propose improvements in existing algorithms involved in the operation of autonomous vehicles. The brain and its functions were analyzed from the cognitive psychology point of view [8]. In special, the cognitive processes related to memory, in its different levels.

The ability of storing information in memory and recalling it when necessary is fundamental to allow the execution of procedures, and the pursuit of long-term goals. As important as the capacity of remembering concepts and experiences, is the ability of forgetting what is irrelevant and focusing attention to what momentarily matters.

We analyzed the similarities between the memory architecture believed to exist in the brain and the process of building maps in robotics. Inspired by these similarities, we proposed a novel strategy to remove noise from occupancy grid maps called map decay. Map decay consists in merging sensory information obtained in runtime (online) with *a priori* data from a high-precision map constructed offline. Online or offline data are emphasized according to whether the map cells are observed by the sensors or not. Cells observed by sensors are updated using traditional occupancy grid mapping techniques. However, cells that are not observed are adjusted so that their occupancy probabilities tend to the values found in the offline map after some time. The effect of this adjustment is an apparent fading (decay) of online information in unobservable regions of the map, while high precision offline information is retained.

*This study was financed in part by Coordenação de Aperfeiçoamento de Pessoal de Nível Superior – Brasil (CAPES) – Finance Code 001; Conselho Nacional de Desenvolvimento Científico e Tecnológico - Brasil (CNPq); and Fundação de Amparo à Pesquisa do Espírito Santo - Brasil (FAPES) – grant 594/2018 and FAPES/VALE 509/2016.

Thomas Teixeira, Vinicius Brito, Lucas Veronese, Claudine Badue, Thiago Oliveira-Santos, and Alberto F. De Souza are with the Departamento de Informática, Universidade Federal do Espírito Santo, Vitoria, ES, 29075-910, Brazil (phone: +55-27-4009-2138; fax: +55-27-4009-5848; e-mail: {thomas, alberto}@lcad.inf.ufes.br).

Filipe Mutz is with the Coordeação de Informática, Instituto Federal do Espírito Santo, Serra, ES, 29173-087, Brazil (email: filipe.mutz@ifes.edu.br).

The main objective of using map decay is to correct occupancy probabilities of map cells that were improperly[1] adjusted because of moving objects or incorrect sensor measurements. Although further observations of the cells would correct their occupancy probabilities, due to several factors the cells may not be observed again. Among these factors, we can mention the sparsity of sensors, the presence of sensorial blind spots, and the natural movement of the autonomous car away from the cell.

Map decay addresses this issue by turning the cells probabilities into the values found in the offline map. This strategy was developed under the assumption that the most precise information available about an unobservable cell is the value present in the high-precision offline map. Map decay was successfully employed and still in use in the IARA autonomous vehicle (IARA is an acronym for Intelligent and Autonomous Robotic Automobile, Fig. 1) from Universidade Federal do Espírito Santo (UFES), Brazil.

## II. Memories and Maps

In this section, we review the brain's memory architecture as suggested by cognitive psychology. Posteriorly, IARA's map organization is described and, finally, the analogy between the brain's memory architecture and the mapping processes of IARA is elucidated.

### A. *The Brain's Memory Architecture*

Cognitive psychologists organize memories in three levels: the sensory memory, the short-term memory, and long-term memory [8],[9]. Although all of these memories have the common role of retaining information, they vary in several aspects such as how long they last, how they are encoded, and how much associations they have with other memories.

The sensory memory is the first level of memory, and it stores information coming from our senses for short periods of time. This information is constantly updated and quickly decay as new stimuli arrive [10]. Sensory memory manifests itself as the image that we still see for a moment after closing our eyes, or the sensation of still hearing a loud sound even when the source moves away.

The short-term memory receives information from several sources (both external information from sensors, and internal information from brain processes) and it is able to store integrated information for long enough to enable the achievement of simple day-by-day tasks. It also has control processes that allow the exchange of information with the long term memory, which increases the capacity of keeping information [8], [10].

The long-term memory can store large amounts of information for long periods. The information stored in this memory, such as names of old friends, places where we lived, and what we have learned, can be brought to short-term memory if necessary and used in tasks [10]. The long-term memory is the ground over which our life experience is build. It stores the network of concepts that allows us to make sense of the world and of our sensorial data, and is the brain's most consolidated information.

### B. *IARA's Mapping Architecture*

IARA represents the world using occupancy grid maps. In these maps, the environment is represented as a two-dimensional grid. Each cell of the grid stores the probability that the region it covers is occupied by obstacles [11], [12]. IARA uses three occupancy grid maps, an instantaneous map, an offline map, and an online map.

The instantaneous map is constructed projecting sensorial data into a momentary occupancy grid map. This map only lasts while novel sensory information does not arrive. As soon as new sensor data are received, the last instantaneous map is discarded and a new one is constructed. This map reflects the area of the world observed by the most recent sensor reading. The instantaneous map is not directly used and it is only used as an intermediary resource for the construction of the other maps.

The offline map is constructed prior to runtime by (i) driving IARA over a path while storing its sensorial data, (ii) estimating the poses visited by IARA using as much resources and data as necessary, and (iii) integrating consecutive instantaneous maps in a single map of the whole environment. A human expert performs a post processing in this map to remove imperfection left in the map. The offline map is used in runtime to estimate IARA's position in the world and also as material to the production of the online map.

The online map represents the state of the world in runtime. It is constructed by merging the offline map and the instantaneous map. The motivation to merge these maps is to use the high quality information from the offline map and the recent information from instantaneous maps. The online map is used by the software modules related to navigation to plan the IARA's next actions.

Fig. 2 (a), (b), and (c) show examples of instantaneous, offline, and online map, respectively. In these maps, grayscale pixels represent the cell's occupancy probabilities with black being maximal occupancy probability and white being minimal occupancy probability. Blue pixels represent cells that were not observed by sensors yet.

### C. *Analogy between Memories and Maps*

Like the long-term memory that is responsible for keeping our life long experience, the offline map stores the most consolidated and reliable information the robot has about the world. Such information serves as working material to other processes when the instantaneous information they need cannot be observed by sensors.

Like the short-term memory that integrates information of the long-term memory and other external sensor, the online map integrates consolidated data (offline map, long-term memory) with new data perceived by the sensors and keeps such information as long as necessary to allow the execution of tasks. These tasks take in consideration what IARA learned during the offline phase and also the current appearance of the world, as observed by sensors.

[1] The term improperly is used in the sense that the cell probabilities will not represent the static structure of the environment.

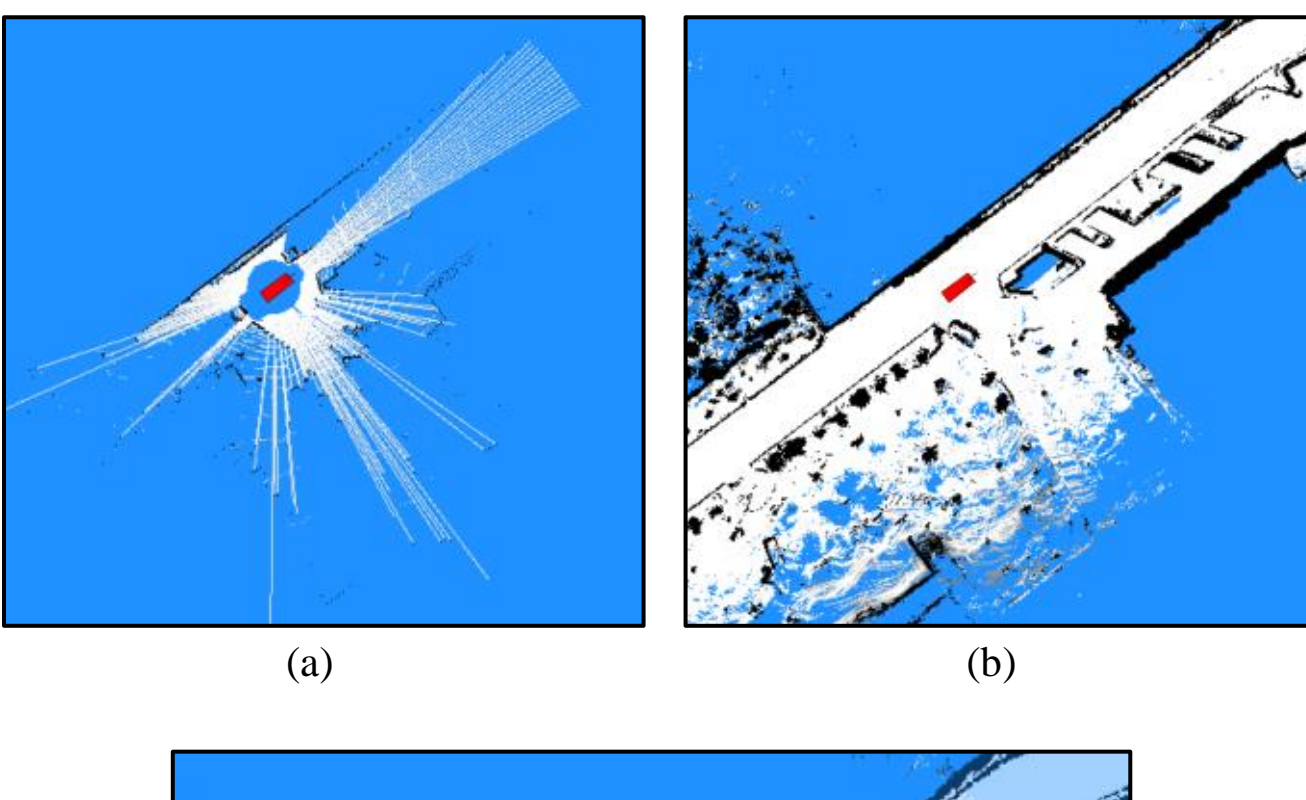

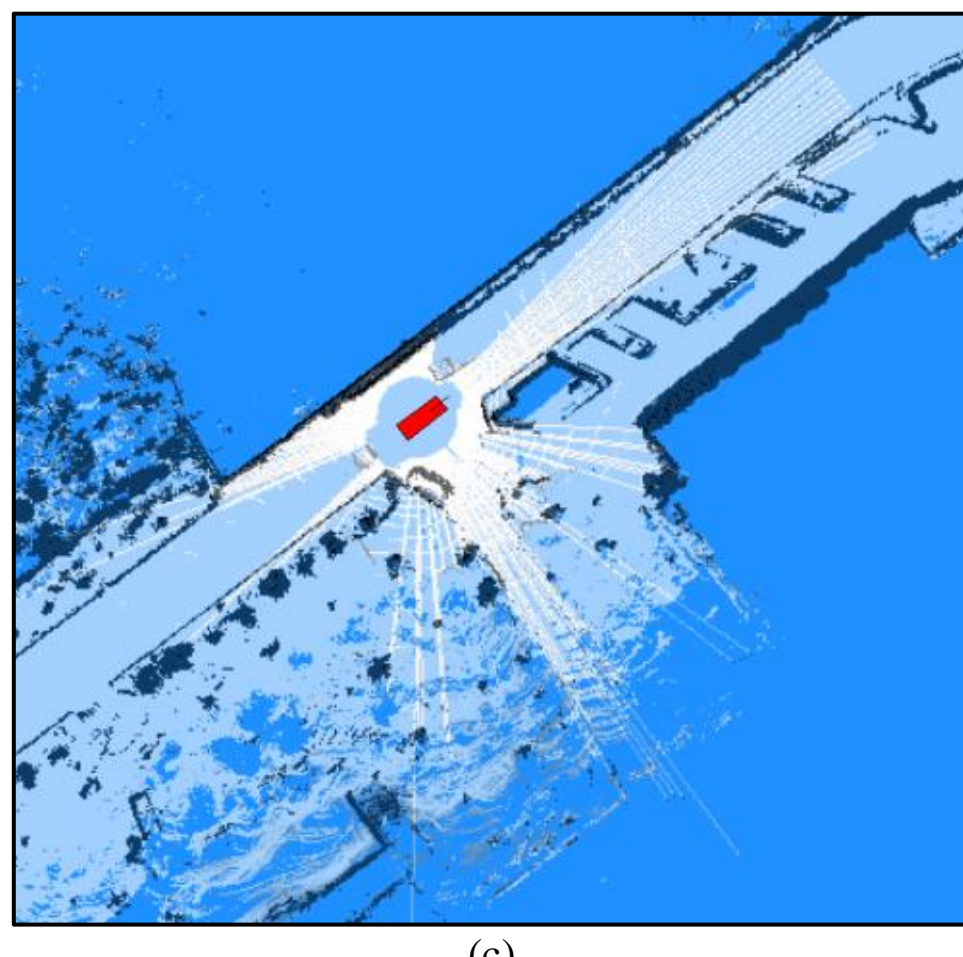

Figure 2. Visualization of the maps used by IARA. Blue pixels were not observed by sensors yet, and pixels in shades of gray represet the occupancy probability of the cells. Black pixels represent maximal occupancy probability, and white represent minimal occupancy probability (a) The instantaneous map that is constructed using a single Velodyne point cloud, and represents the sensorial memory. (b) The offline map that stores the most consolidated information IARA has about the world, and represents the long-term memory. (c) The online map that is used posteriorly by the navigation algorithms, and represents the short-term memory.

To complete our metaphor, the instantaneous map, like the sensory memory, keeps the information captured by sensors, while a new reading is not performed. However, in this work, differently from the human sensorial memory in which information fades away with time, we consider that the instantaneous map decays immediately with the arrival of new information.

The analogy between maps and memories has the potential to inspire insights and new algorithms. This potential is demonstrated by the map decay strategy. Map decay was inspired by the brain's ability of releasing (forgetting) from shot-term memory information that is no longer necessary and the ability of using information from the long-term memory to make sense of incomplete sensorial information.

## III. IARA'S MAPPING ARCHITECTURE

In this section, we present the hardware and software of IARA, with emphasis to the set of modules that support its mapping system. Following, the algorithm to construct the offline and instantaneous maps from sensorial data is presented. Finally, the algorithm for creating the online map and the map decay strategy are explored.

### *A. The IARA Platform*

IARA's hardware is based on a Ford Escape Hybrid, which was adapted by Torc Robotics[2] to enable (i) electronic actuation, of the steering wheel, throttle and brake; (ii) reading internal information (e.g., car odometry); and (iii) powering several high-performance sensors and computers. IARA is equipped with the following set of sensors: one Light Detection and Ranging (LIDAR) Velodyne HDL 32-E; a high precision GPS RTK Trimble; four stereo cameras Point Grey Bumblebee; an Inertial Measurement Unit (IMU) XSENS MTi; and up to four computers Dell Precision R5500 (currently two are in use).

IARA's software modules were structured using the LCAD's version of CARMEN framework [13]. This local version is open to the scientific community and can be accessed in https://github.com/LCAD-UFES/carmen_lcad. The software modules are factored into four main functions: mapping, localization, navigation, and control. Mapping addresses the problem of continuously creating a map of the environment, which contains information describing the places the car may or may not be able to navigate through. Localization addresses the problem of estimating the car's pose (position and orientation) relative to the origin of the map. Navigation addresses the problem of continuously planning a trajectory (list of control commands composed of velocity and steering wheel angle, along with the respective execution durations) from a car's state to a goal state using the map. Finally, the control module is responsible for calculating the set of steering wheel, throttle and brake efforts that allows the car to follow the trajectory planned by the navigation module. Other auxiliary modules were developed to provide additional functionalities such as obstacle avoidance, health monitoring, behavior selection, logging, and simulation. The Mapping Architecture

We will refer to the set of modules related to mapping as the mapping architecture. The sensors used by the mapping architecture are GPS, IMU, car odometry (linear velocity, and steering wheel angle), and the Velodyne 3D point clouds. The mapping architecture presents two different operation modes depending on whether it is running offline or online.

In the offline mode, a human drives IARA over a path and the data captured by the sensors are stored in a log file. The poses visited by IARA are calculated using the pose-based GraphSLAM presented in [12]. GraphSLAM is a Full-SLAM algorithm that uses all the data collected by sensors to estimate the robot poses. The set of estimated poses along with the Velodyne data are sent to the mapping system in order to construct the offline map.

[2] http://www.torcrobotics.com

In the online mode, IARA drives itself autonomously from a starting point to a destination goal. In this operation mode, only the most recent data are used to compute IARA's pose and to update the map. The process to calculate the pose in online mode consists of two steps. In the first step, odometry, IMU and GPS data are fused together using a particle filter [14]. This data fusion process is called fused odometry and it outputs a 6D pose (*x*, *y*, *z*, *roll*, *pitch*, *yaw*). In the second step, the *x*, *y*, and *yaw* components of the pose estimated by fused odometry are refined by a Monte Carlo localization algorithm (the values of the other components remain the same) [11]. This algorithm refines the values of the components using a second particle filter. Firstly, particles are spread using odometry data and IARA's motion model. After that, the obstacles detected by Velodyne are projected to 2D using the technique presented in [12] and compared with the offline map to weight the particles. Finally, particles are resampled and the average pose is returned. As in the offline mode, the pose estimated by the localization algorithm, and the Velodyne point cloud are sent to the mapping system in order to construct the online map.

The mapping system, as the whole mapping architecture, has two modes operation modes, an online and an offline. In both modes, the occupancy probabilities of the cells are updated considering whether Velodyne rays hit obstacles or not. This update follows the traditional occupancy grid map algorithm [11]. Both, the online and offline modes, also has unique features that differentiates them.

The online mode has two particularities not present in the offline mode. The first difference is that the cells of the online map are initialized with the values found in the offline map. By doing so, the initial values of the online map are the most consolidated information available about the environment. As soon as new information arrives, the cells are updated to reflect the actual state of the world. The second difference of the online map is that the map is subject to map decay before the update performed by the occupancy grid mapping algorithm. Given enough time, map decay will correct imperfections left in cells unobservable by sensors.

The offline mode also has a peculiarity that is not present in the online mode. In order to improve the quality of the offline map, a post-processing is performed by a human expert. In this post-processing, the expert cleans potential imperfections existent in the map, e.g., traces left by from moving objects, or incorrect obstacles caused by sensorial noise.

In the following subsections, we present the general occupancy grid mapping rule used to update the map probabilities in both modes of operation and, then, we describe the map decay algorithm used only in online mode.

### B. *Occupancy Grid Mapping*

The maps maintained by IARA are updated for every new point cloud captured by the Velodyne. The point cloud is firstly moved from the sensor reference system to the world reference system using a pose-dependent transform. After that, an instantaneous map is constructed using the point cloud. Finally, the cells of the map (either online or offline) are updated according to the occupancy probabilities of the instantaneous map.

The instantaneous maps construction is directly related to the principle of data acquisition of the Velodyne. The sensor revolves a set of 32 lasers horizontally. The 32 lasers are positioned in an array with increasing vertical angles (the first laser points downwards, the second points a bit higher than the first, and so on). We will refer to this set of 32 rays as a vertical scan. A Velodyne point cloud is composed of several vertical scans captured with different horizontal angles.

The instantaneous map is constructed (i) by calculating the probability that each ray of a vertical scan hit an obstacle, (ii) by projecting the 3D points hit by the rays to the floor (i.e. to 2D), and (iii) by adjusting the occupancy probabilities of the cells corresponding to the projected point accordingly. The strategy to calculate the probability that a ray hit an obstacle is described in [12] and an alternative strategy is found in [15]. The occupancy probabilities of all IARA's maps are represented using log-odds [11]. To update the long-term maps (either online or offline), the cells of the instantaneous map are projected to the respective cells of the long-term maps and the log-odds (that represent the occupancy probability) are summed.

Besides adjusting the cells hit by the Velodyne rays, we also perform a raycast and set to *free* (i.e. to zero probability) all cells between the position of the first ray of a vertical scan and the position of the first ray that hit an obstacle in the same vertical scan. These cells are set to *free* because we assume that if there was an obstacle between these points, some ray would have hit it.

### C. *Map Decay*

Map decay is a new method to remove imperfections from occupancy grid maps using as inspiration the brain's abilities of releasing information that is no longer necessary from short-term memory and of making sense of incomplete sensorial data by filling it with long-term knowledge. These imperfections have several causes. When a dynamic object crosses the cells of a map, for example their occupancy probabilities are raised. Due to IARA's motion, these cells may no longer be observed leading to a trace in the map. The same happens when a false obstacle is detected due to the natural sensorial randomness. As before, the posterior observation of the region should correct the occupancy probability. However, if the cells are not observed either because the robot is moving, or because the cells are into a sensorial blind spot, the occupancy probability will not be corrected.

Imperfections left by moving objects could potentially be handled by one of the several techniques proposed in the literature for mapping in dynamic environments [12],[16]. The conventional technique to handle dynamic objects is to detect and track these objects, and then either treat them as outliers (in landmark maps) or filter them out from the map (in occupancy grid maps) [17] [18][19]. A second technique is to store in the map an history of how the cells change over time, and try to identify moving obstacles analyzing the patterns of changes in the spatiotemporal data [20] [21]. Map decay is a simpler and more efficient (in terms of use of computational resources) solution than the previously listed methods.

Map decay strategy consists of making the probabilities of the cells of the online map tend to the values of the offline map with time. All cells of the online map are updated according to the following rule:

$$M_{on}(x,y) = \frac{M_{on}(x,y)\,W_{on} + M_{off}(x,y)\,W_{off}}{W_{on} + W_{off}} \qquad (1)$$

where $M_{on}(x,y)$ refers to the cell with coordinates (x, y) from the online map, $M_{off}(x,y)$ refers to the cell with the same coordinates from the offline map, $W_{on}$ and $W_{off}$ are importance weights defined by an expert and used to adjust how fast the decay will happen. Preliminary experiments have shown that $W_{on} = 10$ and $W_{off} = 1$ present a good tradeoff between the maintenance of obstacles recently observed and the removal of long-term imperfections of the map.

The map decay equation (Equation 1) can be seen as a weighted average between the values of the offline map and the values of the online map. The effect of map decay is that imperfections left in the map fade away when they are not observed by sensors for some time. The values of the unobservable cells of the online map are slowly replaced by the values found in the offline map. This last statement justifies the importance of having a good offline map. It is important to highlight that the effects of map decay are only relevant to map cells not observed by sensors for some time. Regions observed by sensors, either free or occupied, retain their values and are not significantly affected by map decay sensors.

## IV. Experiments and Results

### A. Datasets

To validate the mapping system, we used two real-world datasets collected in a large-scale and complex environment, the campus of the Universidade Federal do Espírito Santo (UFES, Federal University of Espirito Santo). The first dataset was used to build the offline map and the second to build the online map and to evaluate the map decay strategy. To highlight how advantageous it is to use map decay, the online dataset was recorded at a time with large amount of traffic.

### B. Experimental Evaluation

The online mapping system with map decay was evaluated qualitatively. Fig. 3 presents a comparison between the maps constructed by the online mapping system with and without map decay. Fig. 3 (a) and Fig. 3 (b) show the Velodyne point cloud (blue dots) and the occupied cells of the map (red boxes) when a car is overtaking IARA. It can be seen that traces are left in the map due to the moving car. Because the cells affected by the car's motion fall into a sensor blind spot, the traces are not cleaned up posteriorly.

The remaining sequences of figures compare the map in presence and absence of map decay. Fig. 3 (c), (e) and (g) show that without map decay, the cells of the map remain occupied even when the car already passed IARA. As noted previously, the explanation for this issue comes from the fact that Velodyne rays do not observe the affected cells again once they fall into a blind spot of the sensor. The inability to clean the cells hinders and can even forbid the execution of IARA's navigation plan. Fig. 3 (d), (f) and (h) show the effect of using map decay in the same situation. As time goes by, the cells set as occupied fade away, the information captured by sensors online decay, and their occupancy probabilities becomes the values found in the offline map. It is important to highlight that map cells observed by the sensor and occupied by obstacles still showing a high occupancy probability. Such situation is exemplified by the car behind IARA. Even when map decay is used, the car stills being represented in the map. A video that demonstrate the map decay strategy in action can be accessed in https://youtu.be/cyIfGupar-8 . In the video, the reader can appreciate the behavior of the algorithm with different decay rates.

## V. Conclusions and Future Work

In this work, a relation between the memory architecture believed to exist in the human brain and the mapping algorithms of an autonomous car is presented. The offline map, a high precision representation of the world constructed in an offline phase is associated to the long-term memory. The snapshot map, a map constructed using a single sensor reading is associated with the sensorial memory. Finally, the online map given by the integration of the offline and snapshot maps and used by the navigation module to plan the actions of the autonomous vehicle is associated to the short-term memory.

This analogy inspired a new strategy to remove noise from occupancy grid maps called map decay. Map decay consists of making the occupancy probabilities of the online map tend to the values of the offline map as time passes. It allows the correction of the values of map cells modified improperly and not observed again by the sensors. If not properly handled such situations could forbid the correct operation of the autonomous vehicle.

An important assumption of map decay is the high quality of the offline map. Although the structural quality of the map is guaranteed by the GraphSLAM algorithm, the occupancy probability of their cells still subject to noise left by moving objects and incorrect measurements. In this work, these imperfections in the map were corrected by a human expert. In future works, the possibility of performing these corrections autonomously will be studied. In special, the removal of noise left moving objects will receive special attention.

## Acknowledgements

This study was financed in part by Coordenação de Aperfeiçoamento de Pessoal de Nível Superior – Brasil (CAPES) – Finance Code 001; Conselho Nacional de Desenvolvimento Científico e Tecnológico - Brasil (CNPq); and Fundação de Amparo à Pesquisa do Espírito Santo - Brasil (FAPES) – grant 594/2018 and FAPES/VALE 509/2016.

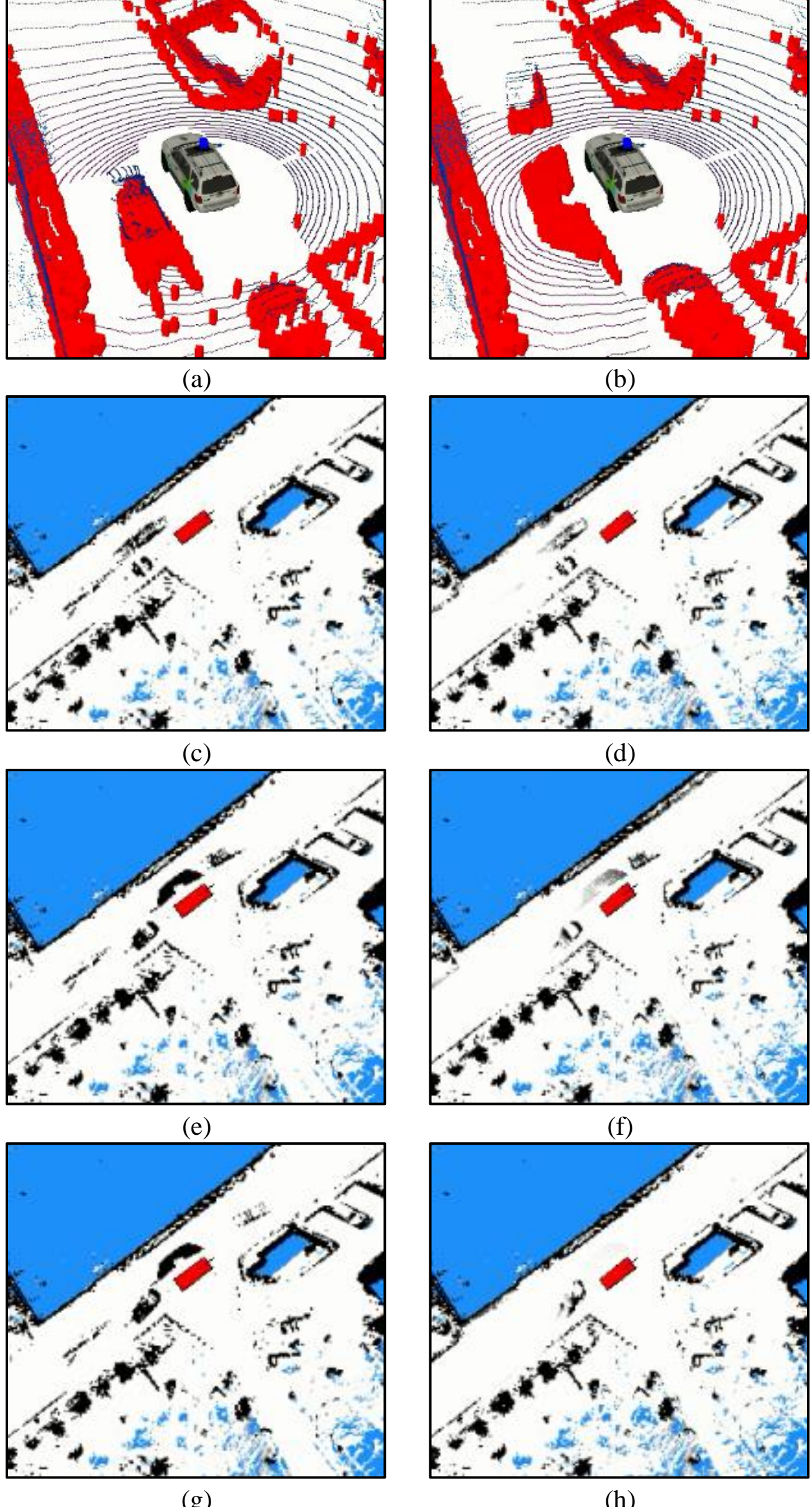

Figure 3. Comparison of the online mapping system output with and without map decay in an overtaking situation. Figures (a) and (b) show a 3D visualization of the overtaking. In these figures, blue dots represent the points of a Velodyne reading and red boxes represent the cells of the with high occupancy probability. The following figures shows IARA's online map in the same situation. If map decay is not used, cells to the left of the car are marked as obstacles and they are not set as free again because they fall into a sensor blind spot (see sequence of figures (c), (e) and (g)). If map decay is employed, however, the cells marked as obstacles slowly fade away as they decay to the offline map values (see sequence if figures (d), (f), and (h)).